\documentclass[10pt,twocolumn,letterpaper]{article}

\usepackage{wacv}              

\usepackage{graphicx}
\usepackage{amsmath}
\usepackage{amssymb}
\usepackage{booktabs}

\usepackage{algorithm}
\usepackage[noend]{algpseudocode}
\usepackage{multirow}
\usepackage{tabularx}
\usepackage[flushleft]{threeparttable}
\usepackage{makecell}
\usepackage{dblfloatfix}

\usepackage[pagebackref,breaklinks,colorlinks]{hyperref}

\usepackage[capitalize]{cleveref}
\crefname{section}{Sec.}{Secs.}
\Crefname{section}{Section}{Sections}
\Crefname{table}{Table}{Tables}
\crefname{table}{Tab.}{Tabs.}

\def\wacvPaperID{796} 
\def\confName{WACV}
\def\confYear{2024}

\begin{document}

\title{Patch-based Selection and Refinement for Early Object Detection}

\author{Tianyi Zhang\\
University of Minnesota\\
{\tt\small zhan9167@umn.edu}
\and
Kishore Kasichainula\\
Arizona State University\\
{\tt\small kkasicha@asu.edu}
\and
Yaoxin Zhuo\\
Arizona State University\\
{\tt\small yzhuo6@asu.edu}
\and
Baoxin Li\\
Arizona State University\\
{\tt\small baoxin.li@asu.edu}
\and
Jae-Sun Seo\\
Cornell University\\
{\tt\small js3528@cornell.edu}
\and
Yu Cao\\
University of Minnesota\\
{\tt\small yucao@umn.edu}
}
\maketitle

\begin{abstract}
   

   Early object detection (OD) is a crucial task for the safety of many dynamic systems. Current OD algorithms have limited success for small objects at a long distance. To improve the accuracy and efficiency of such a task, we propose a novel set of algorithms that divide the image into patches, select patches with objects at various scales, elaborate the details of a small object, and detect it as early as possible. Our approach is built upon a transformer-based network and integrates the diffusion model to improve the detection accuracy. As demonstrated on BDD100K, our algorithms enhance the mAP for small objects from 1.03 to 8.93, and reduce the data volume in computation by more than 77\%. The source code is available at \href{https://github.com/destiny301/dpr}{https://github.com/destiny301/dpr}
\end{abstract}

\section{Introduction}\label{sec:intro}
\begin{figure}[t]
  \centering
   \includegraphics[width=\linewidth]{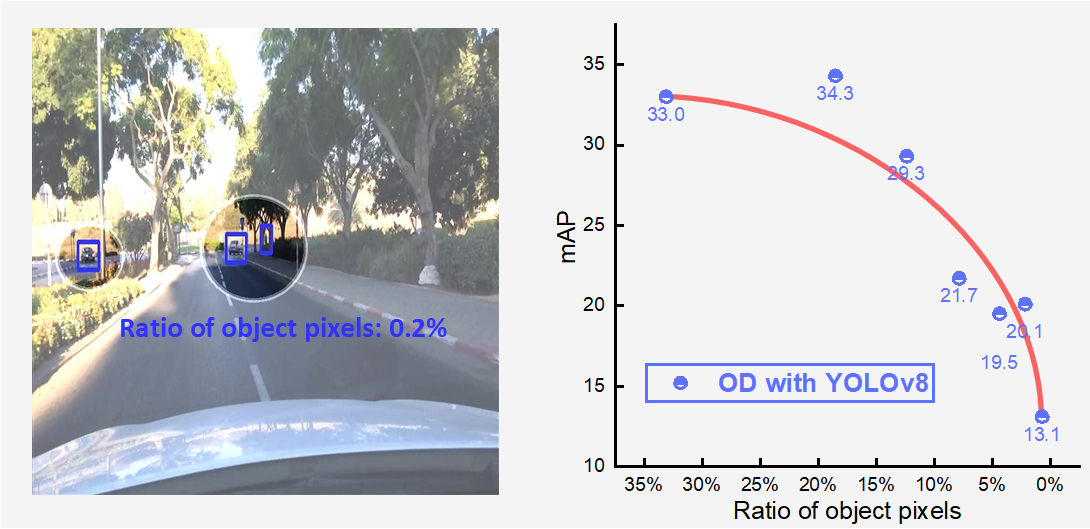}

   \caption{(Left): Objects occupy only a small proportion of the entire image in this example of BDD100K dataset. (Right): With object pixels decreasing, the OD performance rapidly drops.}
   \label{fig:intro}
\end{figure}

Object detection (OD) plays a vital role in numerous real-world applications, such as autonomous driving, and robotics. 
Despite the proliferation of diverse algorithms for this task, existing methods still face significant challenges in early object detection, a crucial aspect enabling prompt and proactive decision-making. In such scenarios, objects in captured images are often significantly reduced in size due to long distances. As illustrated in \cref{fig:intro}, when images contain only a limited number of objects, and the performance of object detection significantly deteriorates due to insufficient data volume.



To address this challenge, we can exploit super-resolution (SR) algorithms to reconstruct the higher-resolution images, thereby augmenting the data available for subsequent object detection models. SR is also a classic problem in computer vision, boasting a plethora of solutions tailored for this task.
Recently, the diffusion models, such as DDPM\cite{ho2020denoising}, have showcased remarkable capabilities in image generation and demonstrated greater stability compared to generative adversarial networks (GAN)\cite{goodfellow2020generative}. Moreover, research focusing on the application of Conditional Diffusion Models (CDM)\cite{saharia2022image, ho2022cascaded}, for SR  has yielded notable advancements. 
Through the utilization of diffusion models for high-resolution image generation, we can achieve substantial enhancements in object detection performance, particularly for datasets with a low object-to-image ratio. However, diffusion models come with a significant computational cost, which poses a challenge for real-world applications like autonomous driving. From the image example in \cref{fig:intro}, the holistic refinement of the image results in a considerable computational burden on background pixels, leading to an excessive waste of resources that does not yield any meaningful contributions to OD.

In this paper, we introduce a novel algorithm, named Dichotomized Patch Refinement (DPR), to tackle the aforementioned problem. DPR leverages CDM to exclusively reconstruct patches that encompass objects, employing a Patch-Selector module for accurate patch classification. While the task of directly localizing small objects presents considerable challenges, discerning the presence or absence of objects within patches proves to be a more feasible approach. By leveraging the Patch-Selector module, we can efficiently filter out irrelevant patches that do not contribute to the subsequent OD task. This strategy significantly reduces the data volume, enabling the immediate generation of refined images using CDM to greatly enhance object detection accuracy. To facilitate the module's implementation, we devise a hierarchical patch encoder inspired by the structure of the Swin Transformer\cite{liu2021swin} to extract embeddings for individual patches. Furthermore, we incorporate a patch classifier through the introduction of a classification token, following a similar approach to ViT\cite{dosovitskiy2020image}. Moreover, in line with our network's hierarchical structure, we introduce a pyramid patch class label to ensure an ample inclusion of positive patches. Our experiments, conducted on the BDD100K dataset, provide compelling evidence of DPR's efficacy and accuracy for early object detection.

To summarize, our key contributions are as follows:
\begin{itemize}
    \item We design a Patch-Selector module, incorporating the attention mechanism, to effectively sift desired patches containing objects from images. Moreover, we introduce a hierarchical architecture and employ a pyramid loss function to further improve the selection process.
    \item By harnessing the capabilities of Conditional Diffusion Models (CDM), we effectively refine solely the selected patches, yielding enhanced performance in object detection.
    \item By enlarging negative patches with interpolation, we seamlessly combine all processed patches to form complete images. Through comprehensive experiments on both patches and entire images, we demonstrate that our DPR achieves competitive early object detection performance with 77.2\% reduction of the computation.
\end{itemize}


\section{Related work}\label{sec:related_work}
\subsection{Diffusion Models for Image SR}
Initially, ConvNets gained prominence in image super-resolution\cite{mao2016image, kim2016accurate}, particularly with the seminal work on the SRCNN model\cite{dong2014learning}.
However, the introduction of generative adversarial networks (GAN) by Goodfellow \etal~\cite{goodfellow2020generative} revolutionized the field, offering unprecedented image generation capabilities.
GAN-based SR methods\cite{wang2018esrgan, ledig2017photo, karras2019style, karras2017progressive, chan2021glean}, have since become prevalent. These techniques employ game theory-inspired competition between a generator and a discriminator to drive iterative improvements and generate high-quality images. Nonetheless, challenges related to training stability and model convergence persist in GAN-based SR methods.

Instead, diffusion models\cite{sohl2015deep} have demonstrated superior performance in image generation and exhibit enhanced stability. The introduction of DDPM by Jonathan\etal~\cite{ho2020denoising}, has further popularized the use of diffusion models in the field of image generation, displacing the reliance on GANs. 
Additionally, recent research has focused on techniques for fast sampling\cite{phung2023wavelet, nichol2021improved, zheng2022fast, huang2022prodiff, zhang2022fast, hang2023efficient}.
DDIM\cite{song2020denoising} accelerates the sampling process by $10\times$ to $50\times$ through the introduction of a more efficient class of implicit probabilistic models.
Given the remarkable performance of diffusion models in image generation, several studies have explored their application in SR by leveraging CDM. For instance, Saharia \etal proposed SR3 \cite{saharia2022image}, which demonstrates improved SR performance based on CDM. Similarly, Jonathan \etal introduced Cascaded Diffusion Models \cite{ho2022cascaded}, which further advances the field of SR.

\subsection{Object Detection (OD)}
Traditional methods for OD, such as Faster RCNN\cite{girshick2015fast}, primarily rely on convolutional layers. The introduction of anchor boxes in Faster R-CNN, a two-stage OD algorithm, has significantly transformed conventional methodologies. Consequently, numerous convolution-based methods, such as YOLO\cite{redmon2016you, redmon2017yolo9000, redmon2018yolov3, bochkovskiy2020yolov4}, Mask R-CNN\cite{he2017mask}, have emerged and continually improved performance in OD. 

Furthermore, the attention mechanism\cite{vaswani2017attention}, initially introduced in ViT \cite{dosovitskiy2020image} for image classification, has been widely adopted in various computer vision tasks, including OD. This is primarily due to the transformer's ability to model long-range dependencies. Carion \etal proposed DETR \cite{carion2020end}, which formulates OD as a direct set prediction problem and employs a transformer encoder-decoder network. DINO \cite{caron2021emerging}, introduced by Caron \etal, leverages self-supervised learning to develop a new transformer network based on ViT. To reduce computation, Liu \etal proposed Swin Transformer \cite{liu2021swin, liu2022swin}, which incorporates a novel window-based self-attention mechanism. 
Inspired by BERT\cite{devlin2018bert} in natural language processing, Bao \etal presented BEiT \cite{bao2021beit} for computer vision applications. 


\begin{figure*}[t]
  \centering
   \includegraphics[width=\linewidth]{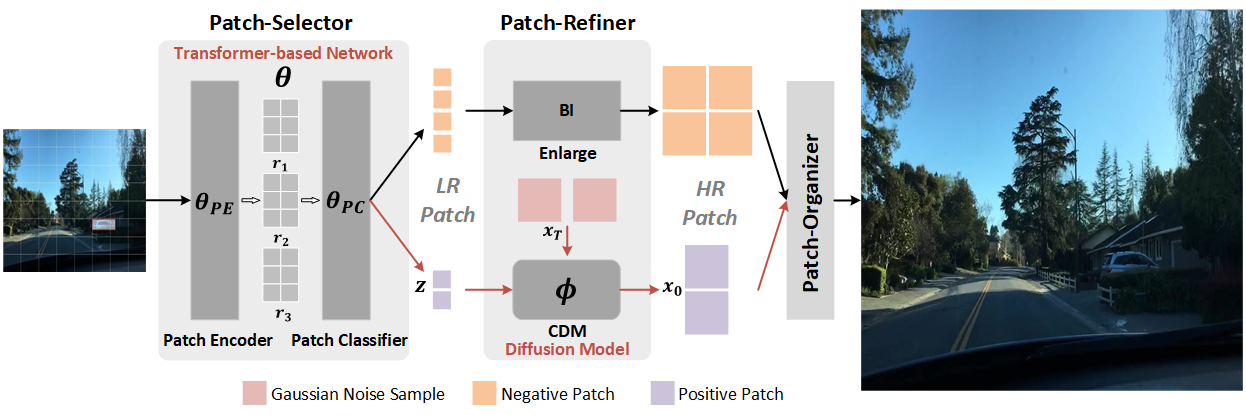}

   \caption{\textbf{Overall architecture of DPR (Dichotomized Patch Refinement).} By dividing all patches of the original image into two groups based on whether it contains objects or not before the image reconstruction, we leverage CDM to process only positive patches to reduce computation and improve the performance for the subsequent OD task since negative patches don't contribute to OD. There are two major components for training: Patch-Selector module with learnable parameters $\theta$, and CDM with parameters $\phi$.}
   \label{fig:flowchart}
\end{figure*}

\section{Methodology} \label{sec:method}
As illustrated in \cref{fig:flowchart}, DPR comprises three crucial modules: Patch-Selector, Patch-Refiner, and Patch-Organizer. The Patch-Selector module is responsible for extracting patch features and performing classification. Subsequently, the Patch-Refiner module elaborates on the positive patches, leveraging CDM to reconstruct them to a higher resolution, thereby enhancing object detection precision. Lastly, to completely show the efficiency and accuracy of our proposed method, we employ inexpensive interpolation techniques to enlarge the negative patches and organize all patches into entire images to facilitate a direct comparison with the original images. In this section, we provide a detailed discussion of all the modules, and outline the specific training procedures of DPR, which are presented in \cref{alg:algo1}. Additionally, Algorithm~\ref{alg:algo2} elucidates the sampling and testing processes.

\begin{algorithm}
\caption{Training with DPR}\label{alg:algo1}
\begin{algorithmic}[1]
\Statex \textbf{Input:} Image data $\boldsymbol{I}_{in}\in\mathbb{R}^{H_{in}\times W_{in}\times C_{in}}$, patch class labels $\boldsymbol{y}_1\in\mathbb{R}^{\frac{H}{2}\times \frac{W}{2}\times 1}$, $\boldsymbol{y}_2\in\mathbb{R}^{\frac{H}{4}\times \frac{W}{4}\times 1}$, $\boldsymbol{y}_3\in\mathbb{R}^{\frac{H}{8}\times \frac{W}{8}\times 1}$, and hyper-parameters $\alpha_{1:T}$
\State Randomly initialize Patch-Selector model parameters $\theta$, and CDM model paramters $\phi$
\For{each epoch t = 1, 2, ...}
\State $\boldsymbol{r}_{1}, \boldsymbol{r}_2, \boldsymbol{r}_3 = f_{\theta_{PE}}(\boldsymbol{I}_{in})$ \Comment{Patch encoding with TL}
\State $\boldsymbol{s}_i = softmax(f_{\theta_{PC}}(\boldsymbol{r}_i))\;\;\; \forall i\in {1, 2, 3}$ \Comment{Patch classification}
\State $\mathcal{L}_{P}(\theta) = \sum_{i=1}^3(-\boldsymbol{y}_ilog(\boldsymbol{s}_i) - \beta (1-\boldsymbol{y}_i)log(1-\boldsymbol{s}_i))$  \Comment{Pyramid loss}
\State $\theta \leftarrow \theta - \eta_\theta \nabla_\theta \mathcal{L}_P(\theta)$  \Comment{Update Patch-Selector model}
\EndFor
\State \textbf{end for}
\State $\boldsymbol{s} = max(\boldsymbol{s}_1, \boldsymbol{s}_2, \boldsymbol{s}_3)$ \Comment{Aggregation of predictions}
\State $\boldsymbol{z} = \boldsymbol{I}_{in}\times \boldsymbol{s}$ \Comment{Positive patch selection}
\State \textbf{repeat}
\State \;\;\;\;\;\;$(\boldsymbol{z}, \boldsymbol{x}_0)\sim p(\boldsymbol{z}, \boldsymbol{x}_0)$ \Comment{Sample positive patch data}
\State \;\;\;\;\;\;$t\sim Uniform(\{1, \ldots, T\})$
\State \;\;\;\;\;\;$\boldsymbol{\epsilon}\sim \mathcal{N}\mathbf{(0, I)}$
\State \;\;\;\;\;\;$\phi \leftarrow \phi - \eta_\phi  \nabla_\phi  \lVert f_\phi(\boldsymbol{z}, \sqrt{\Bar{\alpha_t}}\boldsymbol{x}_{0} + (1-\Bar{\alpha_t})\boldsymbol{\epsilon}, t) - \boldsymbol{\epsilon} \rVert^2$  \Comment{Update CDM model}
\State \textbf{until} converged


\end{algorithmic}
\end{algorithm}

\subsection{Patch-Selector}

\begin{figure}[ht]
  \centering
   \includegraphics[width=\linewidth]{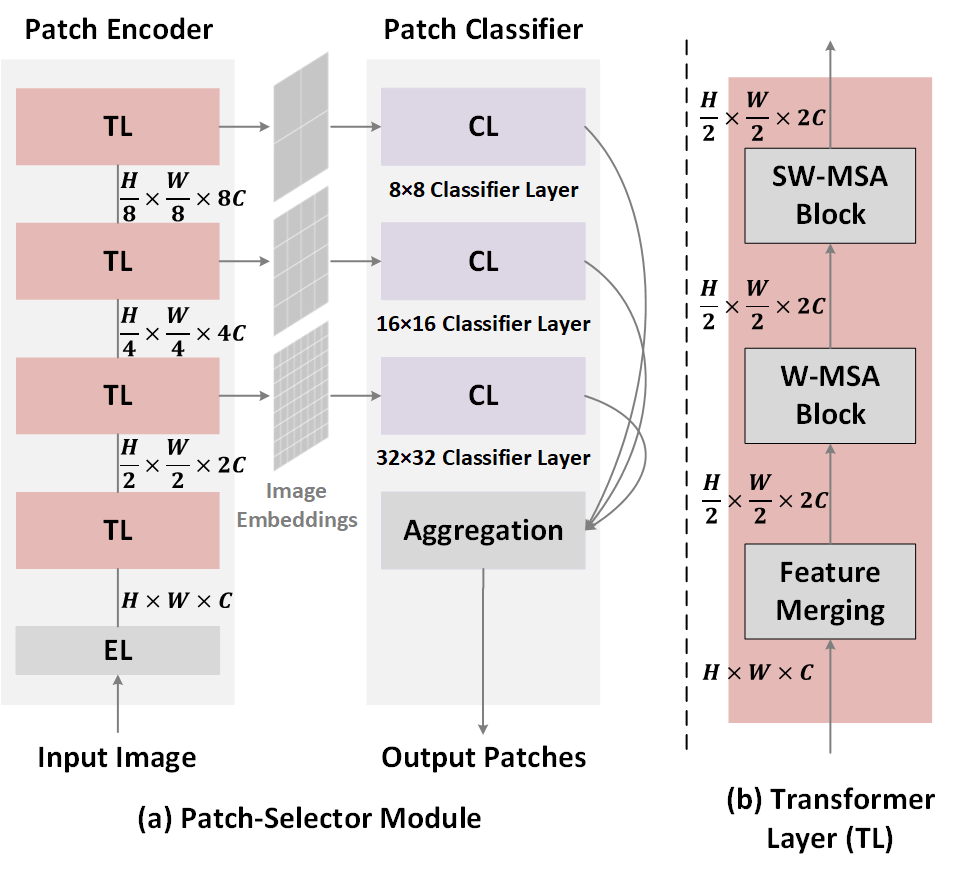}

   \caption{\textbf{The design of Patch-Selector Module.} (a) Utilizing a hierarchical architecture encoder, input images are embedded into features at three different scales. Subsequently, patches within these features undergo classification and aggregation to form the final output. (b) Each Transformer Layer (TL) includes a feature merging block and multiple window-based self-attention blocks.}
   \label{fig:Patch-Selector}
\end{figure}

\textbf{Network architecture.}
This module splits the image into $8\times8$ patches non-overlapping and classifies them to determine if it contains objects or not. Specifically, as depicted in \cref{fig:Patch-Selector}, the input image, $\boldsymbol{I}_{in}\in\mathbb{R}^{H_{in}\times W_{in}\times C_{in}}$ ($H_{in}$, $W_{in}$, and $C_{in}$ are the input image height, width and the number of channels), undergoes a hierarchical patch encoder comprising multiple Transformer Layers (TL). This process generates patch representations at three different scales, namely $\boldsymbol{r}_1\in\mathbb{R}^{\frac{H}{2}\times \frac{W}{2}\times 2C}$, $\boldsymbol{r}_2\in\mathbb{R}^{\frac{H}{4}\times \frac{W}{4}\times 4C}$, $\boldsymbol{r}_3\in\mathbb{R}^{\frac{H}{8}\times \frac{W}{8}\times 8C}$, as the following equations,
\begin{align}
    & \boldsymbol{r}_1 = TL_1(TL_0(EL(\boldsymbol{I}_{in})))
    \\
    & \boldsymbol{r}_2 = TL_2(\boldsymbol{r}_1)
    \\
    & \boldsymbol{r}_3 = TL_3(\boldsymbol{r}_2)
\end{align}
where $TL_i$ denotes the $i$th Transformer Layer, and $EL$ is the embedding layer at the beginning of the network. $H$, $W$ depend on the patch size.

Our TL is similar to the Swin Transformer structure, and it consists of three components: one feature merging layer for representation down-sampling, one window-based multi-head self-attention block (W-MSA), and another shifted window-based multi-head self-attention block (SW-MSA) to capture information across windows. Specifically, W-MSA splits the input feature into $n*n$ non-overlapping windows, where $n$ depends on the window size and feature size, and captures global contextual information within each window. W-MSA solely considers connections within each window, potentially missing out on connections across windows. To address this limitation, SW-MSA shifts the feature by the half of window size before partitioning to enable the cross-window connections.

Specifically, the features $\boldsymbol{r}_1$, $\boldsymbol{r}_2$, and $\boldsymbol{r}_3$ correspond to patches of size $\frac{2H_{in}}{H}\times\frac{2W_{in}}{W}$, $\frac{4H_{in}}{H}\times\frac{4W_{in}}{W}$ and $\frac{8H_{in}}{H}\times\frac{8W_{in}}{W}$, respectively. To classify these patches, we compute the cross-attention with the learnable classification token, denoted as $\boldsymbol{c}$. The computation can be expressed as follows:
\begin{equation}
    \boldsymbol{Q}_i = \boldsymbol{r}_iW_i^q, \;\;\;\boldsymbol{K}_i = \boldsymbol{c}W_i^k, \;\;\;\boldsymbol{V}_i = \boldsymbol{c}W_i^v \;\;\; \forall i\in {1, 2, 3}
\end{equation}
\begin{equation}
    \boldsymbol{A}_i = softmax(\frac{\boldsymbol{Q}_i\boldsymbol{K}_i^T}{\sqrt{d}})\boldsymbol{V}_i \;\;\; \forall i\in {1, 2, 3}
\end{equation}
where $W_i^q$, $W_i^k$, and $W_i^v$ are linear layer weights for query, key, and value matrices.

Next, the features are passed through a multi-layer perceptron (MLP) and a softmax layer to predict the class for each patch as follows,
\begin{equation}
    \boldsymbol{s}_i = softmax(MLP_i(\boldsymbol{A}_i)) \;\;\; \forall i\in {1, 2, 3}
\end{equation}
where $MLP_i$ denotes the output layer for the $i$th feature embeddings.

Accordingly to the network structure, we introduce a pyramid label that contains, $\boldsymbol{y}_1\in\mathbb{R}^{\frac{H}{2}\times \frac{W}{2}\times 1}$, $\boldsymbol{y}_2\in\mathbb{R}^{\frac{H}{4}\times \frac{W}{4}\times 1}$, and $\boldsymbol{y}_3\in\mathbb{R}^{\frac{H}{8}\times \frac{W}{8}\times 1}$, to supervise the training of Patch-Selector module by assigning positive labels to the patches that contain objects.

To minimize information loss, the final prediction is derived by selecting the maximum value from three scales after up-sampling to the same size with bilinear interpolation, ensuring the retention of a greater number of patches.




\begin{algorithm}
\caption{Sampling and Testing with DPR}\label{alg:algo2}
\begin{algorithmic}[1]

\Statex \textbf{Input:} Image data $\boldsymbol{I}_{in}\in\mathbb{R}^{H_{in}\times W_{in}\times C_{in}}$, and hyper-parameters $\alpha_{1:T}$

\State $\boldsymbol{z} = f_\theta(\boldsymbol{I}_{in})$ \Comment{Partition to patches and classify}
\For{each patch $\boldsymbol{z}^1, \boldsymbol{z}^2, ..., \boldsymbol{z}^K$ }
\If{$\boldsymbol{z}$ is positive}
\State Sample $\boldsymbol{x}_T\sim\mathcal{N}\mathbf{(0, I)}$ 
\For{t = T, ..., 1}
\State $\boldsymbol{\epsilon}_t\sim \mathcal{N}\mathbf{(0, I)}$ if $t>1$, else $\boldsymbol{\epsilon}_t=0$
\State $\boldsymbol{x}_{t-1}\leftarrow \frac{1}{\sqrt{\alpha_t}}(\boldsymbol{x}_t-\frac{1-\alpha_t}{\sqrt{1-\Bar{\alpha_t}}}f_\phi(\boldsymbol{z}, \Tilde{\boldsymbol{x}}_t, t)) +\beta_t \boldsymbol{\epsilon}_t$ \Comment{Remove the noise iteratively}
\EndFor
\State \textbf{end for}
\Else
\State $\boldsymbol{x}_0 = Enlarge(\boldsymbol{z})$ \Comment{Enlarge negative patches}
\EndIf
\State \textbf{end if}
\EndFor
\State \textbf{end for}

\State Randomly initialize OD model parameters $\theta^{\prime}$

\For{each epoch t = 1, 2, ...}
\State Update model $\theta^{\prime}$ with $\boldsymbol{x}_0$
\EndFor
\State \textbf{end for}
\State Output the prediction of object classes and bounding boxes with trained model, and evaluate mAP
\end{algorithmic}
\end{algorithm}

\textbf{Loss Function.}
The loss for each patch is computed using cross-entropy. To incorporate predictions from the hierarchical network at three scales and reduce false negative (FN) predictions, we introduce a combined loss formulation. This formulation involves the weighted sum of individual losses and can be expressed as follows
\begin{equation}
    \mathcal{L}_P = \sum_{i=1}^3(-\boldsymbol{y}_ilog(\boldsymbol{s}_i) - \beta(1-\boldsymbol{y}_i)log(1-\boldsymbol{s}_i))
\end{equation}

where $\beta$ is a hyper-parameter to adjust weight, and we set it to 0.01 in our experiments.

\subsection{Patch-Refiner} \label{sec:refine}
Depending on the patch class, different refinement approaches are employed. For positive patches, the conditional diffusion models (CDM) reconstructs them with finer details. Conversely, negative patches are scaled up using simpler up-sampling methods, such as bilinear interpolation (BI), in the Enlarge module.

\textbf{CDM.}
Diffusion Models consist of a forward process that progressively corrupts the input data over $T$ timesteps by keeping adding Gaussian noise, and a reverse process to restore the original data from the final corrupted data. And for CDM, the reconstruction of the corrupted data is performed based on an additional signal that is related to the original data, such as a lower-resolution image in the context of super-resolution (SR).

Let $\boldsymbol{z}\in\mathbb{R}^{H_{p}\times W_{p}\times C_{p}}$ ($H_p$, $W_p$, $C_p$ are the patch height, width, and the number of channels) denote the low-resolution patches we obtain from the Patch-Selector module while $\boldsymbol{x}_0\in\mathbb{R}^{8H_{p}\times 8W_{p}\times C_{p}}$ is high-resolution data. Then, the forward process of our CDM is adding Gaussian noise to $x_0$ over $T$ steps as follows,
\begin{align}
    q(\boldsymbol{x}_t|\boldsymbol{x}_{t-1}) &= \mathcal{N}(\boldsymbol{x}_t; \sqrt{1-\beta_t}\boldsymbol{x}_{t-1}, \beta_t\mathbf{I}) 
    \\
    q(\boldsymbol{x}_{1:T}|\boldsymbol{x}_{0}) &= \prod_{t=1}^T q(\boldsymbol{x}_t|\boldsymbol{x}_{t-1})
    \\
    &=\mathcal{N}(\boldsymbol{x}_t; \sqrt{\Bar{\alpha_t}}\boldsymbol{x}_{0},(1-\Bar{\alpha_t})\mathbf{I})\label{eq:forward}
\end{align}

where $\alpha_{1:T}$, $\beta_{1:T}$ are hyper-parameters, subject to $0<\alpha_t<1$, $\alpha_t+\beta_t=1$, and $\Bar{\alpha_t} = \prod_{i=1}^t\alpha_i$. They determine the variance of the noise added at each iteration. And $\Bar{\alpha_t}$ should be small enough, so that the final signal $x_T$ we acquire after the forward process is roughly also a standard Gaussian noise. 

To gradually recover the original data from the final noise, the CDM model $f_\phi(\boldsymbol{z}, \Tilde{\boldsymbol{x}}_t, t)$ is trained to predict the added noise in each step with the input of low-resolution image $\boldsymbol{z}$, noisy image $\Tilde{\boldsymbol{x}}_t$, and $t$, where the noisy image at timestep $t$ could be obtained from \cref{eq:forward}:
\begin{equation}
    \Tilde{\boldsymbol{x}}_t = \sqrt{\Bar{\alpha_t}}\boldsymbol{x}_{0} + (1-\Bar{\alpha_t})\boldsymbol{\epsilon}, \;\;\;\;\boldsymbol{\epsilon}\sim \mathcal{N}(\mathbf{0}, \mathbf{I}) \label{eq:estimate}
\end{equation}

And for the reverse sampling process, the model recovers the high-resolution patch $x_0$ from $x_T$ conditioned on $z$ with the following equations,
\begin{align}
    &p_\phi(\boldsymbol{x}_{t-1}|\boldsymbol{x}_t, \boldsymbol{z}) = \mathcal{N}(\boldsymbol{x}_{t-1};\mu_\phi(\boldsymbol{z}, \Tilde{\boldsymbol{x}}_t, t), \sigma_t^2\mathbf{I})
\end{align}

We set the variance $\sigma_t^2\mathbf{I}$ to $\beta_t$, and we could compute the mean with the estimated noise from CDM model as follows,
\begin{equation}
    \mu_\phi(\boldsymbol{z}, \Tilde{\boldsymbol{x}}_t, t) = \frac{1}{\sqrt{\alpha_t}}(\boldsymbol{x}_t-\frac{1-\alpha_t}{\sqrt{1-\Bar{\alpha_t}}}f_\phi(\boldsymbol{z}, \Tilde{\boldsymbol{x}}_t, t))
\end{equation}

Finally, the iterative elaboration process is done with the following equation:
\begin{equation}
    \boldsymbol{x}_{t-1}\leftarrow \frac{1}{\sqrt{\alpha_t}}(\boldsymbol{x}_t-\frac{1-\alpha_t}{\sqrt{1-\Bar{\alpha_t}}}f_\phi(\boldsymbol{z}, \Tilde{\boldsymbol{x}}_t, t)) +\beta_t \boldsymbol{\epsilon}_t
\end{equation}
where $\boldsymbol{\epsilon}_t\sim\mathcal{N}(\mathbf{0}, \mathbf{I})$

\textbf{Enlarge.}
In real-world applications, we discard all negative patches since they do not contribute to the subsequent object detection (OD) task. However, this approach can compromise the integrity of the dataset labels, which in turn affects the fairness of experimental comparisons. To ensure a fair evaluation and demonstrate the effectiveness of our approach, we perform scaling on these negative patches using bilinear interpolation (BI), nearest interpolation, or bicubic interpolation, thereby matching them to the same resolution as the positive patches.

\subsection{Patch-Organizer}
By leveraging this module, we combine all the refined positive and negative patches based on their original locations (i.e., the indices of the x-axis and y-axis in the output of Patch-Selector), resulting in the generation of entire images to provide further evidence of the advancements achieved by our DPR algorithm, accompanied by reduced computational requirements.

\section{Experiments} \label{sec:experiment}

\subsection{Dataset and Training Details}
As described in \cref{sec:intro}, we evenly partition the BDD100K dataset\cite{bdd100k} based on the ratio of object pixels into several subsets to test OD, and we select a subset of small ratio to simulate the early detection scenario, where distant objects are typically smaller in size. Our algorithm primarily focuses on enhancing OD performance for the subset with the longest distance, which is named FBDD and consists of images with an object pixel ratio of less than 1.5\%. And we select another subset named NBDD, which contains larger objects with a foreground pixel ratio ranging from 15\% to 23\%, for model fine-tuning. Both subsets contain around 4000 training images and about 1000 validation images with the original size of $1280\times720$.

For Patch-Selector optimization, we resize all the images to be $1024\times1024$ before inputting them to the model. The first embedding layer utilizes a kernel and stride size of 16, with a channel number of 96. We set the learning rates to 0.001 for the convolution-based network and 0.00001 for the attention-based network. For each TL, the depth, window size, and attention head number are set to 2, 7, 3. To align with the hierarchical network structure, we introduce a pyramid label that encompasses three scales: $8\times8$, $16\times16$, and $32\times32$. The patch selection results from the three different scales are then aggregated to a output resolution of $8\times8$.
Once this module gets optimized, input images are resized to be $128\times128$ or $64\times64$ for training.
\begin{table*}[!htb]
    \centering
    \begin{threeparttable}
    \begin{tabularx}{\textwidth}{X|c|c|c|c|c |c|c| c|c|c|c}
    \toprule
    \multirow{2}{*}{\textbf{Patch resolution}} & \multirow{2}{*}{\textbf{Scale}} & \multicolumn{2}{c|}{\textbf{PSNR}$\uparrow$} & \multicolumn{2}{c|}{\textbf{SSIM}$\uparrow$} & \multicolumn{2}{c|}{\textbf{FID}$\downarrow$} & \multicolumn{2}{c|}{\textbf{KID}$\downarrow$} & \multicolumn{2}{c}{\textbf{mAP}$\uparrow$} \\
    \cline{3-12} & & \textbf{BI} & \textbf{CDM} & \textbf{BI} & \textbf{CDM} & \textbf{BI} & \textbf{CDM} & \textbf{BI} & \textbf{CDM} &\textbf{BI} & \textbf{CDM}\\
    \midrule
    $4\times4$ & $\times32$ & 18.06 & 21.64 & 0.7560 & 0.9045 &388.70 &16.32& 0.4624 & 0.0111 & 1.50 & 10.30\\
    $8\times8$ & $\times16$ &  19.96 & 23.76 & 0.8390 & 0.9384 & 276.7 & 8.399& 0.3155 & 0.0033 & 3.48 & 9.14\\
    $16\times16$ & $\times8$ & 22.33&24.85 & 0.9044 & 0.9518 & 161.7 & 8.120 & 0.1707 & 0.0032& 5.89 & 12.00 \\
    $32\times32$ & $\times4$ & 25.61 & 22.00 & 0.9557 & 0.9160 & 51.12 & 23.72 & 0.0440& 0.0147&11.20 & 13.80 \\
    \bottomrule
    \end{tabularx}
    \caption{\textbf{Results of patch refinement.} The patches generated by CDM could provide better features for image classification and OD.}
  \label{tab:CDM results}
    \end{threeparttable}
\end{table*}

We mainly train the CDM to upscale the $16\times16$ patches to $128\times128$ in 1000 timesteps for OD evaluation, although our results show that it also performs well for larger resolution reconstruction. The network architecture is based on U-Net\cite{ronneberger2015u}, with parameters similar to SR3\cite{saharia2022image}. We conduct OD testing using YOLOv8, a state-of-the-art OD algorithm. We experimented with two NVIDIA A6000 GPUs.

\subsection{CDM for Patch Refinement} \label{sec:cdm results}
\begin{figure}
\centering
\begin{subfigure}[!ht]{0.4\textwidth}
    \centering
    \includegraphics[width=\linewidth]{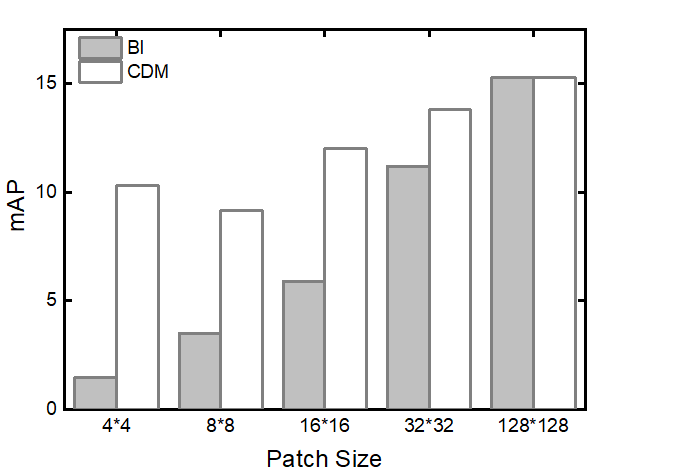}
    \caption[width=0.5\textwidth]{OD results by refining patches of various scales.}
    \label{fig:cdm1}
\end{subfigure}
\hfill
\begin{subfigure}[b]{0.4\textwidth}
    \centering
    \includegraphics[width=\linewidth]{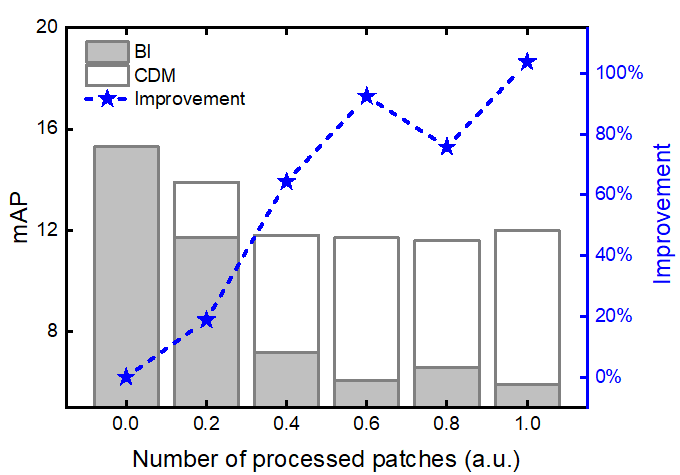}
    \caption[width=0.5\textwidth]{OD results by refining a different number of patches.}
    \label{fig:cdm2}
\end{subfigure}
\caption{(a) CDM performs better for refining images. (b) More processed patches by CDM provide better performance.}
\label{fig:cdm results}
\end{figure}

We perform an extensive evaluation of the CDM for patch refinement, comparing its performance against BI. In \cref{tab:CDM results}, we present the results for 4 different scales with BI or CDM. In all cases, the high-resolution output is $128\times128$ given various input low-resolution patches. 
We evaluate metrics such as Peak Signal-to-Noise Ratio (PSNR) and Structural Similarity Index (SSIM), which measure the similarity between the generated patches and the ground-truth high-resolution patches. Additionally, we employ Fréchet Inception Distance (FID) and Kernel Inception Distance (KID) to compare the features extracted from these patches for image classification. Furthermore, we measure the mean Average Precision (mAP), which serves as the evaluation metric for our primary objective, OD.
Generally, the patches generated by CDM outperform those from BI across all metrics. Despite that $32\times32$ patches from BI exhibit a higher similarity to the original patches, as indicated by the results of PSNR and SSIM, their features for image classification and OD are still inferior to those generated by CDM.  The superiority of CDM is explicitly demonstrated with the mAP comparison in \cref{fig:cdm1}.

In addition, we create some new datasets for OD evaluation by gradually substituting the original high-resolution patches with the processed patches obtained from BI or CDM. As shown in \cref{fig:cdm2}, the OD performance exhibits a more pronounced improvement with an increased proportion of processed patches from CDM. This observation underscores the notable advantages offered by CDM.


\subsection{Patch Selection}
\textbf{Architecture of Patch-Classifier.}
While we have established the viability of CDM for SR, the challenge lies in accurately selecting patches containing objects. Achieving superior performance in the subsequent OD task requires careful consideration of the true positive rate (TPR) during the patch selection stage, as any irreversible information loss at this stage can severely degrade the detection performance. To address this, we utilize multiple transformer layers as the encoder to generate patch embeddings. And our primary focus is on the design of the Patch-Classifier module, which determines the presence of objects in each patch. The impact of the adopted techniques in the design of the Patch-Selector Module is presented in Table \ref{tab:pp results}. 
Initially, we employed convolution layers (Conv-C), which yielded satisfactory results on NBDD dataset. However, by introducing a learnable token and utilizing cross-attention, we achieved even better performance (Attention-C). Moreover, by incorporating a hierarchical network structure and pyramid label (Attention-PC), we observed further improvements across all metrics, particularly in terms of TPR. Comparatively, convolution-based networks also benefited from the hierarchical structure and pyramid label (Conv-PC), but they were unable to match the performance of the attention-based.

\begin{table}[!ht]
    \centering
    \begin{threeparttable}
    \begin{tabularx}{\columnwidth}{p{2.5cm} | c c  c }
    \toprule
    \textbf{Method} & \textbf{TPR (Recall)} & \textbf{maxF} & \textbf{IoU}\\
     \midrule
     Conv-C & 0.7539 & 0.7983 & 0.6350\\
    Attention-C & 0.8700 & 0.8521 & 0.7192\\
    Conv-PC & 0.8277 & 0.8600 & 0.7297\\
    Attention-PC & 0.9084 & \textbf{0.8855} & \textbf{0.7459}\\
    Attention-AC & 0.9511 & 0.8809 & 0.6499 \\
    \textbf{Attention-WC} & \textbf{0.9720} & 0.6946 & 0.6283 \\
    \hline
    FBDD & 0.9101 & - & - \\
    \bottomrule
    \end{tabularx}
    \caption{After adopting all the techniques, our final architecture, Attention-WC, performs the best. The last row is the result of Attention-WC for FBDD dataset while others pertain to NBDD.}
    \label{tab:pp results}
    \end{threeparttable}
    \
\end{table}


\textbf{Aggregation and pyramid loss.}
The results in the sixth row (Attention-AC) of \cref{tab:pp results} demonstrate that incorporating an aggregation block reduces information loss, as evidenced by the higher TPR. Furthermore, by modifying the loss function to place greater emphasis on positive patches, we observed further improvements in TPR, as shown in the seventh row (Attention-WC). With our final Patch-Selector architecture, we achieved a decent TPR for the FBDD dataset, as indicated in the last row of the table.

\textbf{Model size.} 
We explore different model sizes for the Patch-Selector module, specifically using 4, 5, or 6 transformer layers.
In \cref{tab::simplify}, utilizing a network with only 4 transformer layers can achieve equivalent performance in patch selection while reducing FLOPs to 5.01\%.

\begin{table}[!ht]
    \centering
    \begin{threeparttable}
    \begin{tabularx}{\columnwidth}{p{2.32cm} | c c  c }
    \toprule
    \textbf{Method} & \textbf{TPR } & \textbf{\#Params} & \textbf{FLOPs(B)}\\
     \midrule
     Attention-PC/6 & 0.8962 & 1103.99M & 121.47\\
    Attention-PC/5 & 0.8917 & 336.31M & 27.03\\
    Attention-PC/4 & 0.8472 & 119.71M & 6.09\\
    Attention-AC/6 & \textbf{0.9537} & 1103.99M & 121.47\\
    Attention-AC/5 & 0.9459 & 336.31M & 27.03 \\
    \textbf{Attention-AC/4} & 0.9423 & \textbf{119.71M} & \textbf{6.09} \\
    \bottomrule
    \end{tabularx}
    \caption{\textbf{Results for model size.} Four Transformer Layers achieve similar performance with much lower computation.}
    \label{tab::simplify}
    \end{threeparttable}
    
\end{table}



\subsection{Comparison of OD Performance}
To fully demonstrate the merit of our approach, we not only detect objects from patches with bounding box labels different from the original image due to patch partitioning for OD performance comparison, but we also integrate the entire images for detection. As we scale the $16\times16$ patches to $128\times128$, we use the results obtained by directly feeding the $16\times16$ patches into the OD model as the baseline for patch-wise detection. We compare the performance of our DPR with this baseline as well as other methods. Additionally, since we have $8\times8$ patches, the entire image is scaled from $128\times128$ to $1024\times1024$. Similarly, we use the results of the low-resolution $128\times128$ image as the baseline for image-wise detection. 
\begin{table}[!b]
    \centering
    \begin{threeparttable}
    \begin{tabularx}{\columnwidth}{p{1.68cm} | l| c  c c }
    \toprule
    \textbf{Image Size} & \textbf{Method} & \textbf{mAP} & \textbf{TPR} & \textbf{Precision} \\
    \midrule
    $512\times512$ & GT & 7.48 & 0.106 & 0.309\\
    $64\times64$ & GT & 0.194 & 0.017 & 0.009\\
    \hline
    \multirow{5}{*}{$512\times512$} & BI & 0.732 & 0.024 & 0.235\\
    & SwinIR\cite{liang2021swinir} & 0.674 & 0.026 & 0.103 \\
    & SR3\cite{saharia2022image} & 2.38 & 0.061 & 0.423 \\
    & \textbf{DPR}(Ours) & \textbf{4.33} & \textbf{0.078} & \textbf{0.457} \\
    \hline
    $1024\times1024$& \textbf{DPR}(Ours) & \textbf{8.93} & \textbf{0.142} & \textbf{0.274} \\
    \bottomrule
    \end{tabularx}
    \caption{The last row of $1024\times1024$ is upscaled from $128\times128$. DPR performs the best for images-wise OD evaluation.}
  \label{tab:od-image results}
    \end{threeparttable}
    
\end{table}

\begin{table*}[!htb]
    \centering
    \begin{threeparttable}
    \begin{tabularx}{\textwidth}{X | p{1.5cm}|c |c |c |c |c |c |c |c | c | c}
    \toprule
    \multirow{2}{*}{\textbf{Patch Size}} & \multirow{2}{*}{\textbf{Method}} & \multirow{2}{*}{\textbf{\# Patch}} & \multicolumn{2}{c|}{\textbf{mAP}} & \multicolumn{2}{c|}{\textbf{mAP$_{50}$}} & \multicolumn{2}{c|}{\textbf{TPR}} & \multicolumn{2}{c|}{\textbf{Precision}} &\multirow{2}{*}{\textbf{\makecell{FLOPs\\(B)}}}\\
    \cline{4-11} & & & \textbf{PP} & \textbf{AP} & \textbf{PP} & \textbf{AP} & \textbf{PP} & \textbf{AP} & \textbf{PP} & \textbf{AP} & \\
    \midrule
    $16\times16$ & - & - & 1.99 & 1.63 & 4.25 & 3.33 & 0.0493 & 0.0362 & 0.0818 & 0.0677 & -\\
    \hline
    \multirow{4}{*}{$128\times128$} & BI & 100\% & 2.82 & 2.37 & 5.59 & 4.46 & 0.0618 & 0.0382 & 0.2430 & 0.2910 & 34.41 \\
    & SR3\cite{saharia2022image} & 100\% & 4.55 & 3.46 & 8.23 & 6.92 & 0.0843 & \textbf{0.0623} & \textbf{0.3170} & 0.3930 & 34.41\\
    & \textbf{DPR} & \textbf{22.8\%} & \textbf{5.12} & 3.45 & \textbf{9.05} & 6.82 & \textbf{0.0886} & 0.0606 & 0.1930 & 0.3870 & \textbf{7.85}\\
    & \textbf{DPR-B}& - & - & \textbf{3.54} & - &\textbf{7.82} & - & 0.0557 & - & \textbf{0.5200} & -\\
    \bottomrule
    \end{tabularx}
    \caption{\textbf{Results of patch-wise OD.} PP denotes the experiments with only positive patches, and AP is tested for all patches.The third column (\# Patch) shows the ratio of reconstructed patches for each image. mAP is the primary evaluation metric for OD.  Our DPR obtains higher mAP with fewer patches refined.}
  \label{tab:od patch results}
  
    \end{threeparttable}
\end{table*}

\textbf{Patch-wise detection.}
Besides our approach, we generate high-resolution patches with another two methods, BI and SR3\cite{saharia2022image}, for comparison. 
BI simply scales up all patches to $128\times128$ using bilinear interpolation while SR3 is a conditional diffusion model (CDM) based on DDPM that performs entire image super-resolution (SR). 
The PP columns in \cref{tab:od patch results} present the results when we feed only positive patches to OD, which is the approach we adopt in real-world applications. For BI and SR3, we assume that they can perfectly select positive patches (i.e., TPR is 1).
The PP columns show our DPR performs the best. 

\begin{figure*}[!htp]
  \centering
   \includegraphics[width=0.95\linewidth]{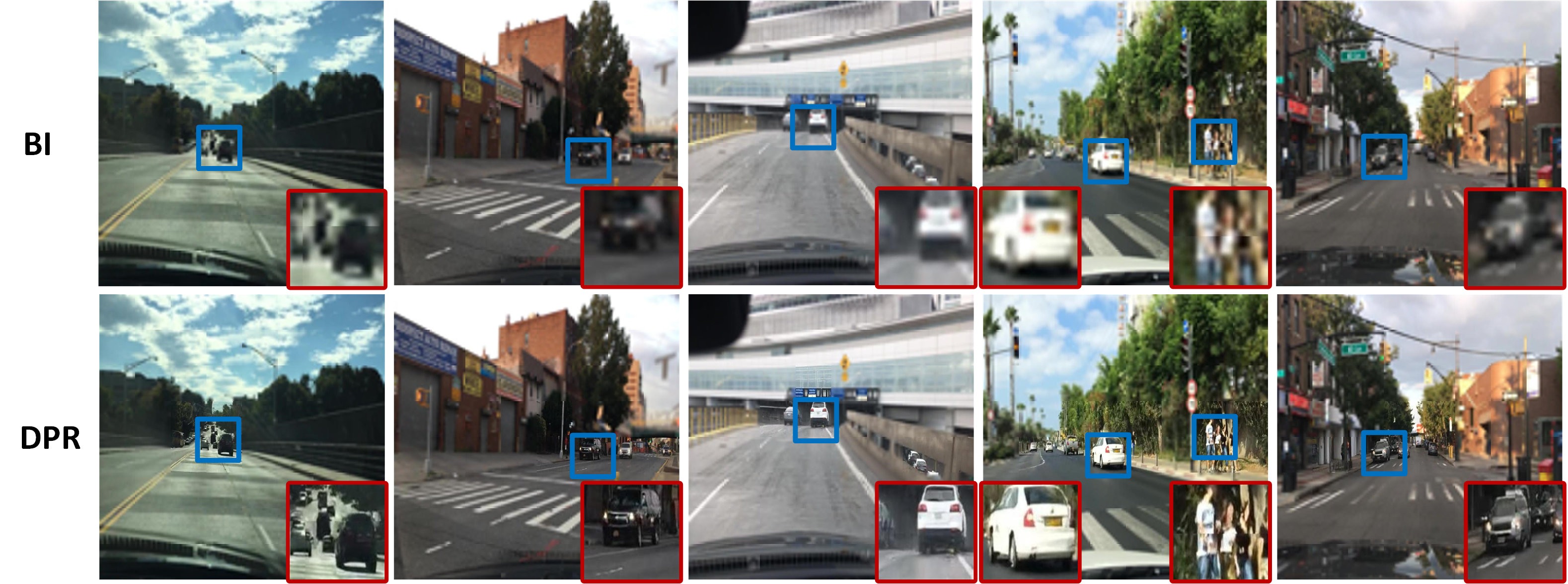}

   \caption{\textbf{The visual comparison of Bilinear interpolation and CDM.} The resolution of all images is $1024\times1024$ generated from $128\times128$ input when each patch of the images is scaled up from $16\times16$ to $128\times128$. Within the red boxes, the enlarged key patches from our DPR that contain objects exhibit finer details.}
   \label{fig:example}
   \vspace{-4mm}
\end{figure*}

To address the potential unfairness in the previous experiments, we also evaluate OD with both positive and negative patches, shown in the AP columns. As mentioned in \cref{sec:method}, negative patches of DPR are enlarged with BI. Additionally, we conduct an experiment where all negative patches are replaced with black patches to simulate the removal of negative patches, denoted as DPR-B. DPR achieves comparable performance to SR3 with significantly fewer average refined patches of each image (14.59 on average versus 64). This highlights the computational efficiency of our approach.
Interestingly, DPR-B outperforms DPR, suggesting that the selection results of our Patch-Selector module contribute to OD. By excluding the negative patches, which may introduce noise and confusion, our approach focuses solely on the positive patches, leading to improved detection results.


\textbf{Image-wise detection.}
\Cref{fig:example} shows the visual comparison of BI and our DPR after integrating patches. While the overall generated images from DPR appear similar to BI, the crucial patches containing objects exhibit finer details, indicating that only a small amount of data need to be processed by CDM, leading to more efficient computation.

Quantitative results for OD are presented in \cref{tab:od-image results}. To compare with another SR method, SwinIR \cite{liang2021swinir}, and maintain consistency, we align our evaluation with SwinIR's setting, upscaling images from $64\times64$ to $512\times512$. We show the results of ground truth high-resolution images in the table as the upper bound. SR3 can perform much better than transformer-based SwinIR.
DPR achieves the highest mAP among all the methods, resulting in an improved mAP from 0.194 to 4.33, while DPR can enhance mAP from 1.03 to 8.93 when upscaling images from $128\times128$ to $1024\times1024$.


\textbf{Efficiency of our approach.}
To trade off the computation and performance, we experiment with various thresholds for patch classification when upscaling images from $64\times64$ to $512\times512$ in \cref{tab:th}. The second row, yielding mAP of 4.33, stands out as the optimal choice, achieving 63\% computation reduction. 

For FBDD up-sampling from $128\times128$ to $1024\times1024$ with the same threshold, our PS module outputs only 22.8\% patches for CDM generation and OD, and the FLOPs of PS are negligible compared to CDM, which means we save 77.2\% computation compared to full-image generation, as demonstrated in \cref{tab:od patch results}.

\begin{table}[!t]
    \centering
    \begin{threeparttable}
    \begin{tabularx}{\columnwidth}{X |c |c c c}
    \toprule
    \textbf{PS TPR} & \textbf{\# Patch} & \textbf{mAP} & \textbf{mAP$_{50}$} & \textbf{Precision}\\
    \midrule
    0.9813 & 71\% & 4.65 & 9.56 & 0.399  \\
    \textbf{0.8972} & \textbf{37\%} & \textbf{4.33} & \textbf{8.96} & \textbf{0.457}  \\
    0.6290 & 11\% & 2.53 & 5.75 & 0.463  \\
    0.2120 & 3\% & 2.53 & 6.12 & 0.422  \\
    \bottomrule
    \end{tabularx}
    \caption{By selecting different thresholds to assign patches, the second row achieves a comparable performance with 63\% computation reduction. PS TPR of 0.8972 means about 90\% positive patches are correctly selected by Patch-Selector (PS).}
  \label{tab:th}
    \end{threeparttable}
    \vspace{-10pt}
\end{table}


\section{Conclusion} \label{sec:conclusion}
In this paper, we propose a novel Dichotomized Patch Refinement algorithm (DPR) to efficiently enhance the OD performance by selectively reconstructing the high-resolution patches of images with conditional diffusion models. With a hierarchical transformer-based network and pyramid loss function, positive patches containing objects are accurately located. With patch-wise CDM, low-resolution positive patches are significantly refined, thereby improving the performance of the subsequent OD task. And the experimental results on the BDD100k dataset show that DPR effectively improves the mAP for early object detection from 1.03 to 8.93 with only 22.8\% computation.

\section*{Acknowledgments}
This work is supported by the Center for the Co-Design of Cognitive Systems (CoCoSys), one of seven centers in Joint University Microelectronics Program 2.0 (JUMP 2.0), a Semiconductor Research Corporation (SRC) program sponsored by the Defense Advanced Research Projects Agency (DARPA).

{\small
\bibliographystyle{ieee_fullname}
\bibliography{egbib}
}

\end{document}